\documentclass[conference]{IEEEtran}
\IEEEoverridecommandlockouts
\renewcommand{\baselinestretch}{0.83}
\usepackage{cite}
\usepackage{amsmath,amssymb,amsfonts}
\usepackage{graphicx}
\usepackage{textcomp}
\usepackage{algpseudocode}
\usepackage{comment}
\usepackage{algorithm}
\usepackage{xcolor}
\usepackage{comment}
\usepackage{url}
\usepackage{hyperref}
\usepackage{eso-pic}
\usepackage{makecell}
\usepackage{subcaption}
\usepackage[compatibility=false]{caption}

\def\BibTeX{{\rm B\kern-.05em{\sc i\kern-.025em b}\kern-.08em
    T\kern-.1667em\lower.7ex\hbox{E}\kern-.125emX}}

\newcommand{\mycopyrightnotice}{\hfill\footnotesize }
\makeatletter
\def\ps@IEEEtitlepagestyle{%
  \def\@oddfoot{\mycopyrightnotice}%
  \def\@evenfoot{}%
}

\makeatother

\title{An AI-IoT Based Smart Wheelchair with Gesture-Controlled Mobility, Deep Learning-Based Obstacle Detection, Multi-Sensor Health Monitoring, and Emergency Alert System}

\author{
\IEEEauthorblockN{
Md. Asiful Islam\IEEEauthorrefmark{1},
Abdul Hasib\IEEEauthorrefmark{2},
Tousif Mahmud Emon\IEEEauthorrefmark{3},\\
Khandaker Tabin Hasan\IEEEauthorrefmark{4},
A. S. M. Ahsanul Sarkar Akib\IEEEauthorrefmark{5}
}

\IEEEauthorblockA{
\IEEEauthorrefmark{1}Department of Computer Science and Engineering,\\
Military Institute of Science and Technology, Bangladesh\\
\IEEEauthorrefmark{2,3}Department of IoT \& Robotics Engineering,\\
University of Frontier Technology, Bangladesh\\
\IEEEauthorrefmark{4}Department of Computer Science,
American International University-Bangladesh\\
\IEEEauthorrefmark{5}Department of Robotics, Robo Tech Valley, Dhaka, Bangladesh
}

\IEEEauthorblockA{
Emails:
\IEEEauthorrefmark{1}asifuli751@gmail.com,
\IEEEauthorrefmark{2}sm.abdulhasib.bd@gmail.com,
\IEEEauthorrefmark{3}1901011@iot.bdu.ac.bd,\\
\IEEEauthorrefmark{4}tabin@aiub.edu,
\IEEEauthorrefmark{5}ahsanulakib@gmail.com
}
}

\begin{document}
\maketitle

\begin{abstract}

The growing number of differently-abled and elderly individuals demands affordable, intelligent wheelchairs that combine safe navigation with health monitoring. Traditional wheelchairs lack dynamic features, and many smart alternatives remain costly, single-modality, and limited in health integration. Motivated by the pressing demand for advanced, personalized, and affordable assistive technologies, we propose a comprehensive AI-IoT based smart wheelchair system that incorporates glove-based gesture control for hands-free navigation, real-time object detection using YOLOv8 with auditory feedback for obstacle avoidance, and ultrasonic for immediate collision avoidance. Vital signs (heart rate, SpO$_2$, ECG, temperature) are continuously monitored, uploaded to ThingSpeak, and trigger email alerts for critical conditions. Built on a modular and low-cost architecture, the gesture control achieved a 95.5\% success rate, ultrasonic obstacle detection reached 94\% accuracy, and YOLOv8-based object detection delivered 91.5\% Precision, 90.2\% Recall, and a 90.8\% F1-score. This integrated, multi-modal approach offers a practical, scalable, and affordable solution, significantly enhancing user autonomy, safety, and independence by bridging the gap between innovative research and real-world deployment.
\end{abstract}


\section{Introduction}
The growing global population of differently-abled and elderly individuals has created a pressing demand for advanced assistive technologies that enhance autonomy and quality of life. According to the World Health Organization (WHO), over 1.3 billion people experience some form of disability, with mobility impairments being particularly common \cite{1}. While wheelchairs serve as essential mobility aids, traditional models lack the dynamic, context-aware features needed for real-world environments\cite{edubot}.

Recent innovations in embedded systems and artificial intelligence have enabled "smart wheelchairs" with gesture control, autonomous navigation, health monitoring, and object recognition \cite{2,2425}. These systems enhance conventional mobility devices with multi-modal interaction and environmental awareness. Gesture-based controllers using accelerometers and gyroscopes enable control without requiring fine motor skills \cite{2}, while deep learning models like YOLOv8 provide real-time obstacle detection and auditory guidance \cite{2425}.

Mobility is crucial for autonomy, employment, and social inclusion, with restricted mobility linked to social exclusion and reduced life satisfaction \cite{akib1}. However, practical deployment of smart wheelchairs is often hindered by high costs, complexity, and lack of modularity. Many solutions remain limited by expensive infrastructure and narrow input modalities that reduce accessibility \cite{5}, with health monitoring often underemphasized in affordable, modular designs. To address these challenges, we propose an integrated Smart Wheelchair System with three key contributions:
\begin{itemize}
    \item Gesture-based mobility control with intelligent obstacle detection: Intuitive hands-free navigation combining gesture control with YOLOv8 object detection and auditory feedback
    \item Continuous multi-sensor health monitoring: Real-time tracking of vital signs (heart rate, SpO$_2$, ECG, temperature) with automated cloud-based alerts via ThingSpeak
    \item Modular low-cost architecture: Customizable, affordable design suitable for resource-constrained environments
\end{itemize}

\section{Literature Review}
Recent advances in smart wheelchairs have explored gesture control, health monitoring, and AI-based obstacle detection. 

For instance, Iqbal et al. propose a joystick-free hand-gesture control scheme using surface electromyography (sEMG) signals and user-independent classifiers, specifically targeting users with finger mobility limitations \cite{6}. Similarly, Newaskar et al. developed a voice-commanded wheelchair system integrating an ESP32 microcontroller, microphone input, and ultrasonic sensors for obstacle avoidance, coupled with health monitoring via pulse oximetry displayed on a mobile app, enabling hands-free control for users with severe impairments \cite{9}. Gesture recognition via computer vision is also prominent; Sadi et al. utilize a CNN-based finger gesture recognition system combined with IoT hardware \cite{10}, while Nguyen et al. employ a YOLOv8n model for real-time gesture classification, achieving minimal latency in directional control \cite{11}. Huda et al. contribute a spatial-relationship-based hand landmark system with simple mathematical thresholds for robust, low-effort gesture interpretation adaptable to diverse conditions \cite{15}. Shivamma et al. offer an affordable Arduino-based solution using MPU6050 sensors and RF modules for head and hand gesture control \cite{19}. In addition to control modalities, several studies focus on integrating physiological and posture monitoring for enhanced safety and health management. Arshad et al. introduce retrofit kits that transform manual wheelchairs into electric ones, featuring multiple operation modes, machine learning-based posture detection, and physiological monitoring aimed at safety and pressure ulcer prevention \cite{8}. Zahid et al.'s PulseRide system dynamically adjusts wheelchair assistance using heart rate and ECG data through reinforcement learning, reducing muscle strain and optimizing physical exertion \cite{16,fall}. 

Khalili and Smyth extend object detection architectures (YOLOv8) to improve small-object detection accuracy, crucial for recognizing subtle navigation obstacles \cite{13}. Other vision-based approaches include Chatzidimitriadis et al.'s deep learning system that tracks head orientation via a single RGB camera for continuous, low-effort wheelchair navigation \cite{17}, and Du et al.'s GC-YOLO model, designed for efficient obstacle detection in wheelchair blind spots \cite{18}. Rahman et al. propose OpenNav, a zero-shot pipeline combining open-vocabulary 2D detection, semantic segmentation, and depth-based masking to generate 3D bounding boxes, enabling flexible navigation without retraining for new object categories \cite{12}. Autonomous and shared control frameworks aim to balance user input with smart assistance. Xu et al. present the CoNav Chair, leveraging the Robot Operating System (ROS) and a shared control navigation algorithm that blends\cite{bionic} user commands with autonomous obstacle avoidance to improve indoor maneuvering and trust \cite{14}. Han et al. focus on posture-based control using dual MEMS inertial sensors and K-means clustering to infer user posture for implicit interaction, enhancing control accuracy and usability \cite{7}.  

These innovations highlight a multidisciplinary effort to create safer, more intuitive wheelchair systems. However, existing solutions often focus on singular functionalities or require costly components. Our system presents an affordable, integrated approach combining gesture-based navigation, real-time YOLOv8 obstacle detection, and IoT-enabled health monitoring, offering a practical alternative that bridges innovation and real-world deployment.

\section{Proposed Methodology}

\subsection{System Architecture and Design Overview} 

\begin{figure}[h]
    \centering
    \includegraphics[width=.9\linewidth, height=4.85cm]{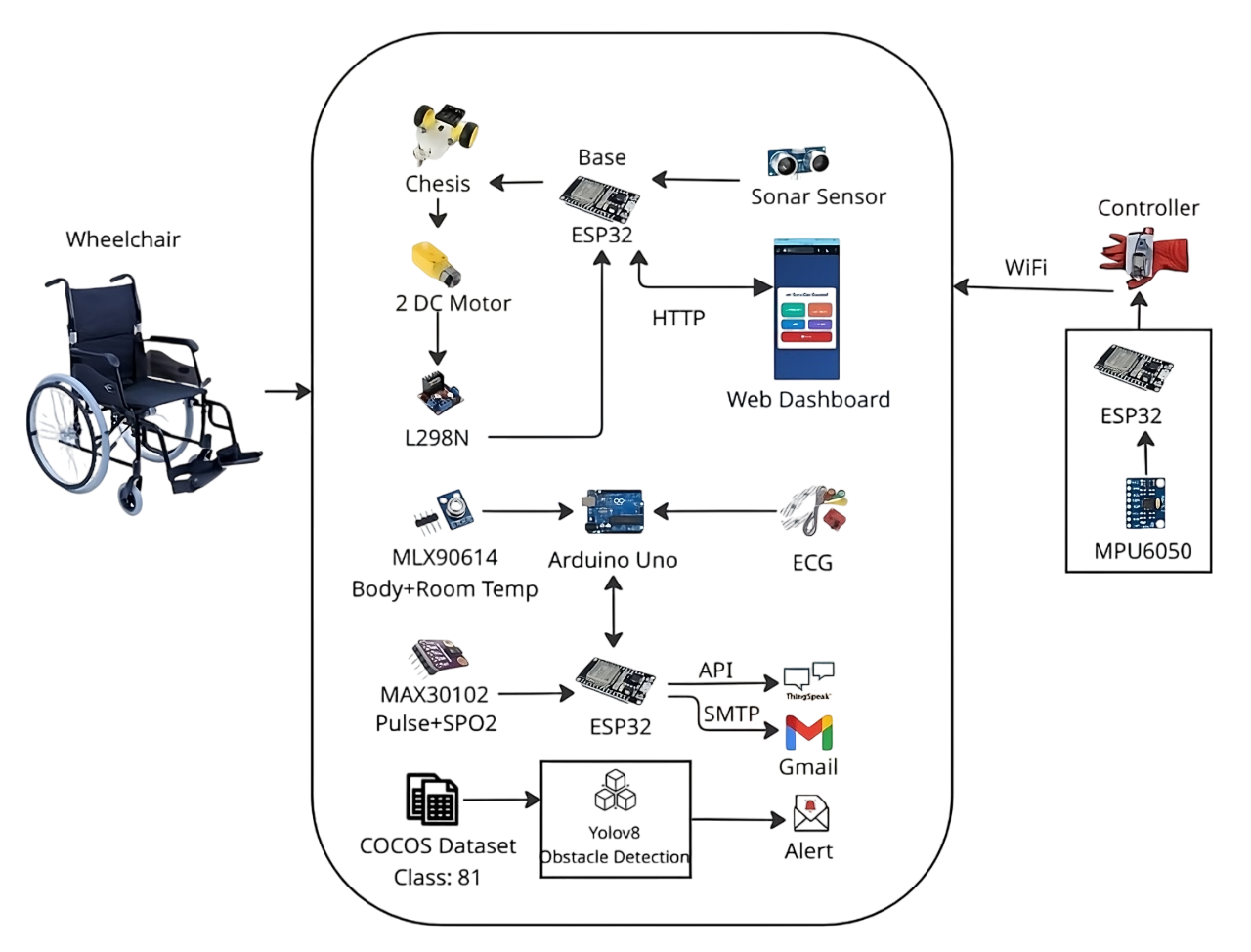}
    \caption{System Architecture of IoT Based Smart Wheelchair}
    \label{fig:sys_arc}
\end{figure}

The Smart Wheelchair System is structured as a multi-component, interconnected framework comprising five distinct modules: the Gesture Controller, the Wheelchair Base, the Health Monitor, the Arduino Sensor Hub, and the Object Detection Module. This distributed architecture, shown in Figure \ref{fig:sys_arc}, enables specialized processing and parallel operation, enhancing the overall system's reliability and responsiveness.
The Gesture Controller interprets the user's hand movements into directional commands, which are wirelessly transmitted to the Wheelchair Base. The Wheelchair Base manages the physical movement of the wheelchair, including propulsion and steering, while performing active obstacle avoidance. The Health Monitor continuously collects vital physiological data from the user, facilitated by the Arduino\cite{akib2} Sensor Hub, which serves as an interface for health sensors. If critical conditions are detected, the Health Monitor triggers alerts and enables remote data access. The Object Detection Module processes real-time video feeds to identify nearby objects and provide auditory feedback, improving situational awareness. Inter-module communication utilizes both wireless and wired protocols, with gesture control signals sent via Wi-Fi-based WebSocket \cite{20} and sensor data exchanged through serial communication. This integrated approach forms a cohesive IoT ecosystem for the smart wheelchair.

\subsection{Working Procedure of the System}

\begin{figure}[h]
    \centering
    \includegraphics[width=.95\linewidth, height=3.35cm]{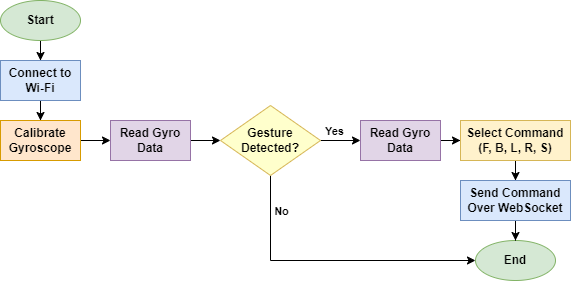}
    \caption{Working Procedure of Wheelchair Base}
    \label{fig:wheelchair_base}
\end{figure}

\begin{figure}[h]
    \centering
    \includegraphics[width=.9\linewidth, height=3.35cm]{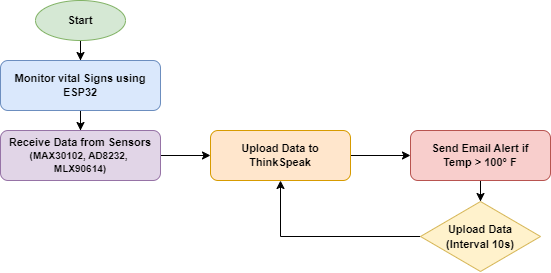}
    \caption{Health Monitoring Procedure}
    \label{fig:hlth_mntr}
\end{figure}

\begin{figure}[h]
    \centering
    \includegraphics[width=.95\linewidth]{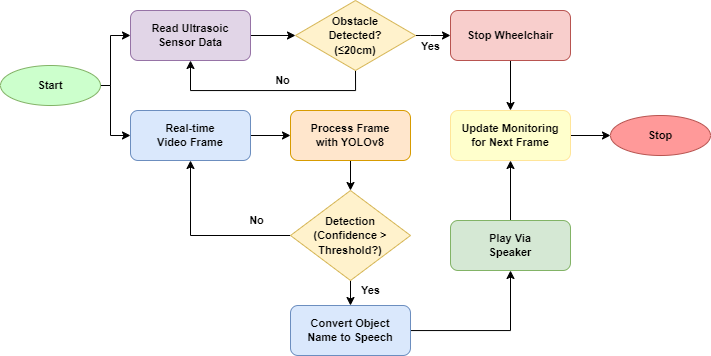}
    \caption{Object Detection Procedure}
    \label{fig:objct_det}
\end{figure}

\begin{figure}[h]
    \centering
    \includegraphics[width=0.87\linewidth]{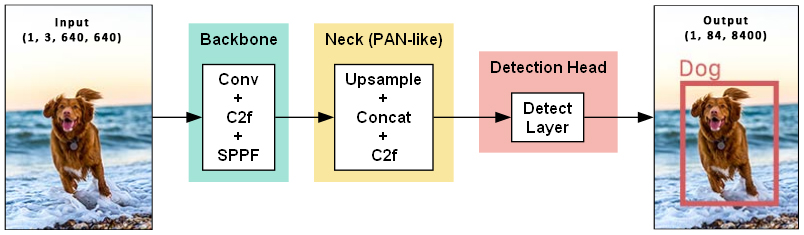} 
    \caption{YOLOv8 Architecture Overview}
    \label{fig:yolov8_arch}
\end{figure}

The Smart Wheelchair System operates through a synchronized and continuous workflow across its interconnected modules, ensuring real-time control, monitoring, and feedback. Upon system activation, the Wheelchair Base module (powered by an ESP32 microcontroller) initiates its Wi-Fi Access Point, creating a local network (e.g., "SmartWheelchair"). It also starts its WebSocket server to listen for incoming control commands and begins continuously monitor for obstacles using its ultrasonic sensor. Concurrently, the Gesture Controller module (built on an ESP32 board) connects to this Wi-Fi network. After a brief gyroscope calibration period, it commences reading the user's hand tilt data from the MPU6050 sensor \cite{21}. Based on the detected gestures, the Gesture Controller generates movement commands (Forward, Backward, Left, Right, Stop) and transmits them wirelessly via WebSocket to the Wheelchair Base. The working procedure of Wheelchair Base is shown in Figure \ref{fig:wheelchair_base}.
In parallel, the Arduino Uno-based Sensor Hub continuously acquires physiological data, reading ECG signals from the AD8232 sensor and temperatures from the MLX90614 sensor \cite{22}. This raw sensor data is then formatted into a CSV string and transmitted via a serial connection to the Health Monitor module. The Health Monitor module processes data from the Sensor Hub alongside heart rate and SpO$_2$ readings from its integrated MAX30102 sensor \cite{25} and runs on an ESP32 board, periodically uploading processed data to the ThingSpeak cloud platform. Figure \ref{fig:hlth_mntr} shows the working flow diagram of this process of monitoring vital signs. Furthermore, the Health Monitor continuously checks the user's body temperature. If it exceeds a predefined threshold (e.g., 100°F), an automated email alert is dispatched to a designated recipient, ensuring timely notification of critical health conditions.
Concurrently, the wheelchair combines ultrasonic and AI-based object detection to enhance navigation and safety, as shown in Figure~\ref{fig:objct_det}. The ultrasonic sensor measures distance and stops movement if an obstacle is within 20 cm, ensuring short-range collision prevention. A camera connected to a Raspberry Pi captures real-time video, processed using OpenCV. The YOLOv8 architecture used for detection is illustrated in Figure~\ref{fig:yolov8_arch}, featuring a Backbone (feature extraction), Neck (feature fusion), and Detection Head (bounding box prediction). This efficient design allows real-time inference on edge devices like the Raspberry Pi. The Ultralytics YOLOv8 \cite{2425} model detects objects with high confidence, while gTTS or pyttsx3 convert detections into spoken alerts. This auditory feedback provides situational awareness beyond basic obstacle avoidance. The AI system extends coverage to objects at varying distances and angles—particularly outside the immediate path—and supports custom training for diverse environments.
\subsection{System Operation and Algorithms}
\subsubsection{Gesture-Controlled Mobility}
The Gesture Controller operates by continuously acquiring data from the MPU6050 gyroscope. Raw angular velocity readings, denoted as \( g_x, g_y, g_z \), are first subjected to a calibration process. This calibration involves collecting a predetermined number of samples, \( N \), while the sensor is held stationary. The average of these samples for each axis is computed to determine the static offsets:
\begingroup
\scriptsize
\[
\text{offset}_x = \frac{1}{N} \sum_{i=1}^{N} g_{x,i}, \quad 
\text{offset}_y = \frac{1}{N} \sum_{i=1}^{N} g_{y,i}, \quad 
\text{offset}_z = \frac{1}{N} \sum_{i=1}^{N} g_{z,i}
\]
\endgroup
These offsets are then subtracted from raw gyroscope readings to obtain calibrated values, which are then compared against predefined thresholds to infer the user's intended movement. The resulting command denoted as \( C \in \{ \text{'F', 'B', 'L', 'R', 'S'} \} \) for Forward, Backward, Left, Right, and Stop, is transmitted to the Wheelchair Base via WebSocket as detailed in Algorithm \ref{alg:gesture_command_algo}.
\begin{algorithm}[h]
\scriptsize
\caption{Gesture Command Generation}
\label{alg:gesture_command_algo}
\begin{algorithmic}[1]
\Procedure{InitializeGestureController}{}
    \State Calibrate MPU6050 sensor
    \State Connect to Wheelchair Base via WebSocket
\EndProcedure
\Procedure{GenerateAndSendGestureCommand}{}
    \While{true}
        \State Read raw gyro data \( (g_x, g_y, g_z) \)
        \State Compute calibrated data \( g'_x = g_x - \text{offset}_x, \quad g'_y = g_y - \text{offset}_y \)
        \State Determine command \( C \):
        \If{ \( g'_y > \text{threshold}_F \) }
            \State \( C \gets 'F' \)
        \ElsIf{ \( g'_y < \text{threshold}_B \) }
            \State \( C \gets 'B' \)
        \ElsIf{ \( g'_x > \text{threshold}_R \) }
            \State \( C \gets 'R' \)
        \ElsIf{ \( g'_x < \text{threshold}_L \) }
            \State \( C \gets 'L' \)
        \Else
            \State \( C \gets 'S' \)
        \EndIf
        \State Transmit \( C \) via WebSocket
        \State Wait 100 ms
    \EndWhile
\EndProcedure
\end{algorithmic}
\end{algorithm}
\subsubsection{Wheelchair Movement and Obstacle Avoidance}
The Wheelchair Base module continuously measures the distance to obstacles using an ultrasonic sensor and simultaneously receives movement commands via WebSocket. If an obstacle is detected within 20 cm, the wheelchair halts immediately, overriding any movement commands. Otherwise, it executes the latest received command to control the motors for forward, backward, left, right, or stop actions. The detailed control and obstacle avoidance procedure is described in Algorithm \ref{alg:wheelchair_control_algo}.
\begin{algorithm}[h]
\caption{Wheelchair Control and Obstacle Avoidance}
\label{alg:wheelchair_control_algo}
\begin{algorithmic}[1]
\Procedure{InitializeWheelchairBase}{}
    \State Configure motor and ultrasonic sensor pins
    \State Start Wi-Fi AP and WebSocket server
    \State Set motors to stop
\EndProcedure
\Procedure{ManageWheelchairMovement}{}
    \While{true}
        \State Receive command $C$; measure distance $D$
        \If{ $D \le 20$ cm }
            \State Stop motors; set last command to 'S'
        \Else
            \State ExecuteMovement($C$)
        \EndIf
        \State Delay 100 ms
    \EndWhile
\EndProcedure
\Procedure{ExecuteMovement}{$C$}
    \State Match $C$: 'F'→Forward, 'B'→Backward, 'L'→Left, 'R'→Right, 'S'→Stop
\EndProcedure
\end{algorithmic}
\end{algorithm}
\subsubsection{Health Monitoring and Alert System}
\begin{algorithm}[h]
\scriptsize
\caption{Health Data Processing and Alerting}
\label{alg:health_monitor_algo}
\begin{algorithmic}[1]
\footnotesize
\Procedure{InitializeHealthMonitor}{}
    \State Configure MAX30102 sensor and serial communication with Arduino
    \State Configure ThingSpeak client and Email client
\EndProcedure

\Procedure{ProcessHealthDataAndAlert}{}
    \While{true}
        \State Read \texttt{irValue} and \texttt{redValue} from MAX30102
        \State Calculate \texttt{beatsPerMinute} and \texttt{spo$_2$}
        \If{serial data is available from Arduino}
            \State Read and parse raw data string to extract \texttt{ecgValue}, \texttt{ambientTempC}, \texttt{objectTempC}, \texttt{leadStatus}
            \State Convert \texttt{objectTempC} to \texttt{objectTempF}
            \If{ \texttt{objectTempF} $\geq$ 100.0 \textbf{and} \texttt{emailSent} == \texttt{false} }
                \State Initiate email alert transmission
                \State Set \texttt{emailSent} = \texttt{true}
            \ElsIf{ \texttt{objectTempF} $<$ 100.0 }
                \State Set \texttt{emailSent} = \texttt{false}
            \EndIf
        \EndIf
        \If{10 seconds have elapsed since last cloud upload}
            \State Upload all current health data to ThingSpeak
        \EndIf
        \State Control update rate to 10 milliseconds
    \EndWhile
\EndProcedure
\end{algorithmic}
\end{algorithm}
The Health Monitor module continuously collects physiological data from the sensors to compute vital metrics like heart rate, SpO$_2$ (via AC/DC ratio), ECG, and temperature. Data from the Arduino Sensor Hub is parsed in real-time, with critical alerts—such as high temperature—triggering automatic email notifications. All readings are periodically uploaded to the cloud for remote monitoring and historical analysis. The full process is outlined in Algorithm~\ref{alg:health_monitor_algo}.
\subsubsection{Real-Time Object Detection and Auditory Feedback}

The Object Detection Module continuously analyzes video frames using a pre-trained YOLOv8 model to identify objects in real time. Detected objects with confidence scores above a predefined threshold are selected, and their class names are converted to speech using a Text-to-Speech (TTS) engine. This auditory feedback enhances the user's situational awareness by announcing nearby objects that may be outside the ultrasonic sensor’s range. 
\subsection{Prototype Implementation}
\begin{figure}[h]
    \centering
    \rotatebox{-90}{
        \includegraphics[height=4.5cm]{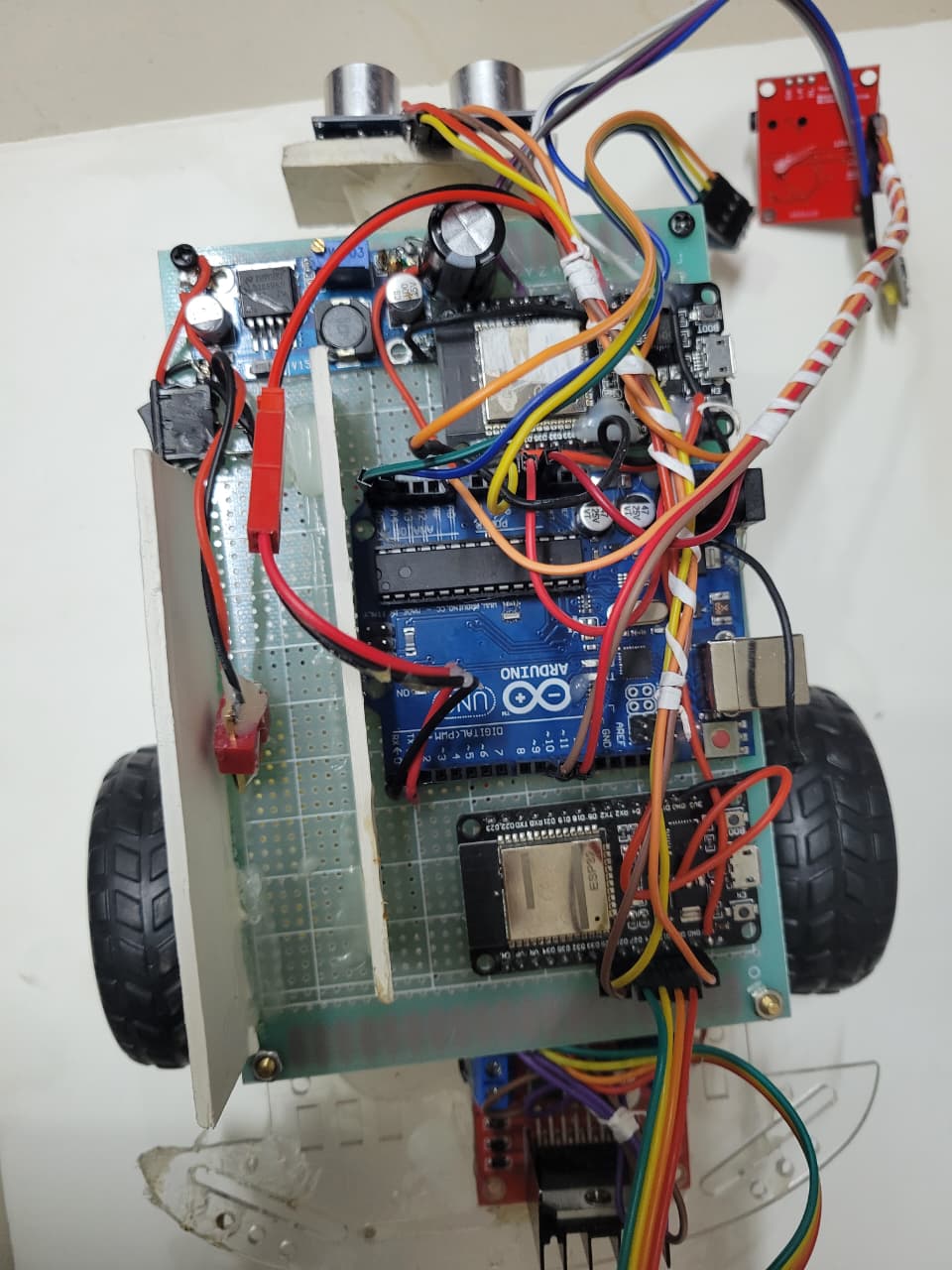}
    }
    \caption{Circuits of Smart Wheelchair system}
    \label{fig:circuit}
\end{figure}

\begin{figure}[h]
    \centering
    \begin{subfigure}[b]{0.42\linewidth}
        \includegraphics[width=\linewidth, height=3.2cm]{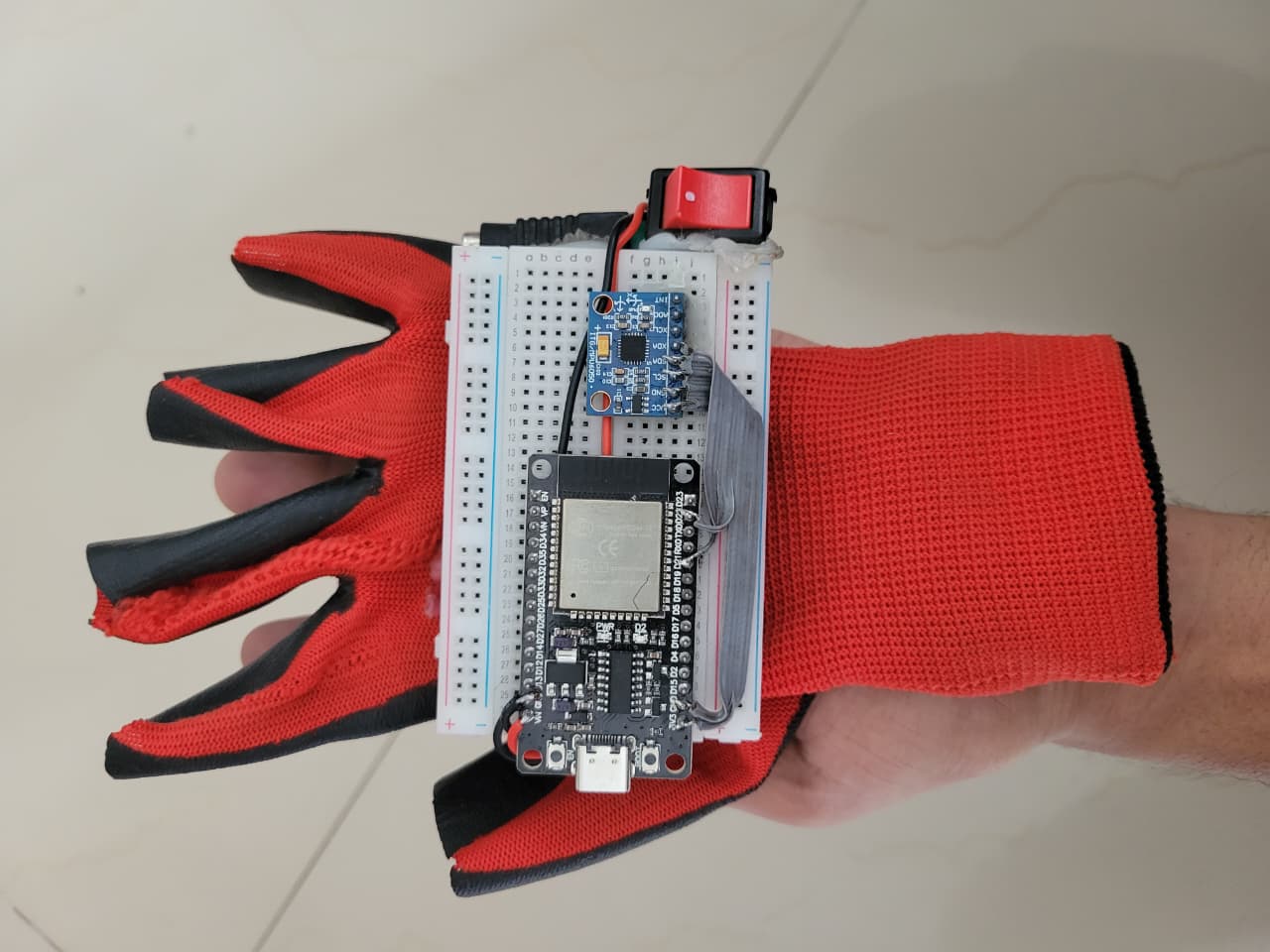}
        \caption{}
        \label{fig:glove_sub}
    \end{subfigure}
    \hfill
    \begin{subfigure}[b]{0.42\linewidth}
        \includegraphics[width=\linewidth, height=3.2cm]{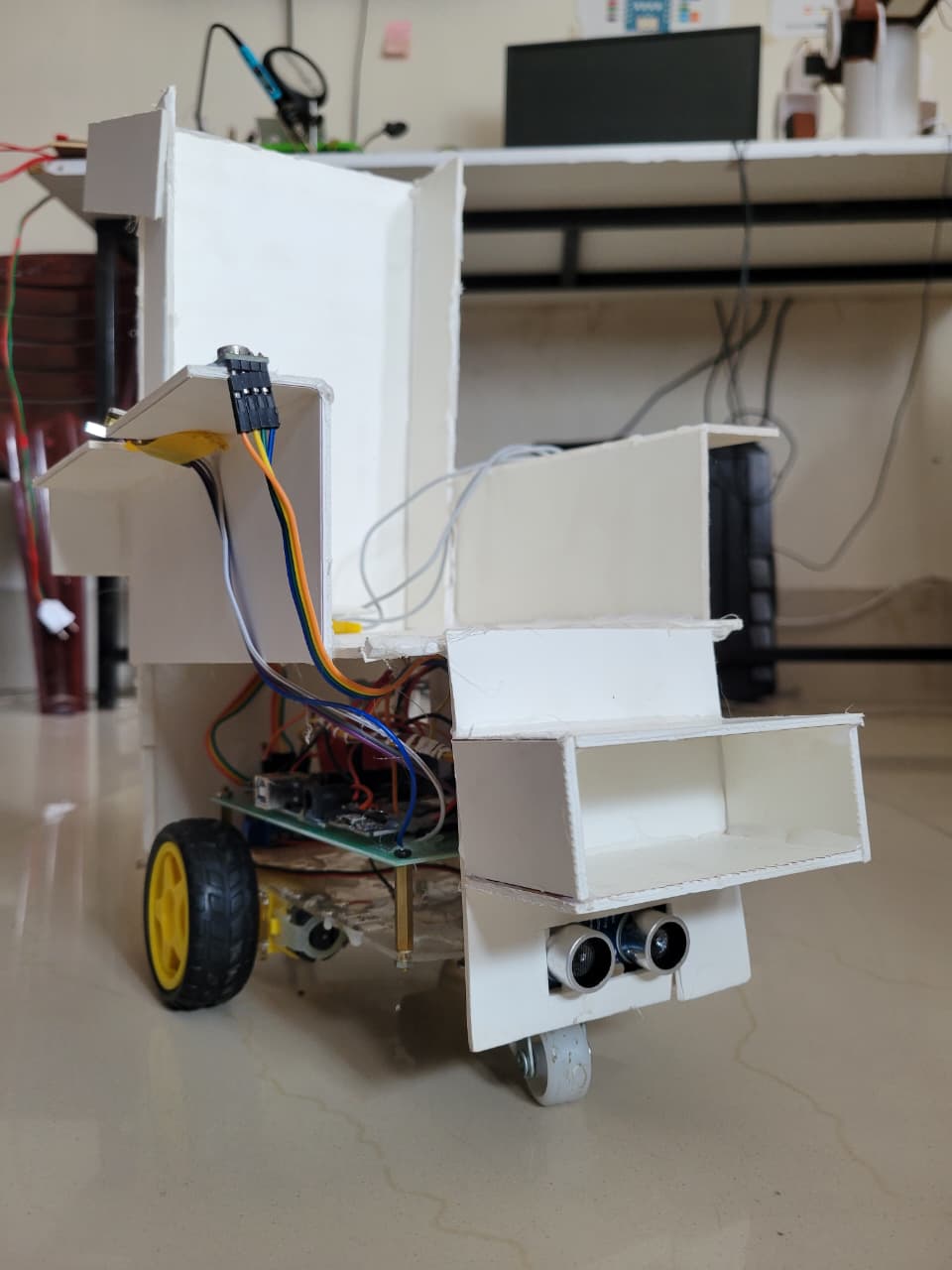}
        \caption{}
        \label{fig:wheelchair_sub}
    \end{subfigure}
    \caption{(a) Hand Glove for Gesture Control, (b) Prototype of the Smart Wheelchair}
    \label{fig:fig5}
\end{figure}

The implementation of the IoT-enabled smart wheelchair integrates gesture-based mobility control, object detection, and health monitoring into a cohesive system aimed at enhancing user autonomy and safety. The internal circuitry of the smart wheelchair system is shown in Figure~\ref{fig:circuit}. The gesture control module is embedded in a hand glove (Figure~\ref{fig:glove_sub}) equipped with an ESP32 and an MPU6050 sensor. After initial calibration, it captures hand tilts and transmits directional commands to the wheelchair over a local Wi-Fi network using the WebSocket protocol.
The wheelchair prototype, shown in Figure~\ref{fig:wheelchair_sub}, is built on a modified frame and integrates an ESP32 receiver, L298N motor driver, and HC-SR04 ultrasonic sensor. This setup enables motor control while ensuring obstacle avoidance through automatic braking if an object is detected within 20 cm.
To further enhance environmental awareness, a Raspberry Pi equipped with a high-definition camera captures forward-facing video and runs an onboard object detection module. The system delivers auditory warnings through a speaker, ensuring users receive alerts for unseen obstacles. Health monitoring is managed by a wearable chest strap that houses ECG, SpO\textsubscript{2}, and temperature sensors, transmitting data securely for local processing and remote access.
\subsection{Implementation Cost}
The prototype's overall cost is roughly \$201.72 (24610 BDT). For an actual wheelchair, an extra \$90 (7500 BDT) will be needed. Table \ref{tbl:cost} presents a detailed breakdown of the equipment and their corresponding prices.
\begin{table}[h]
\scriptsize
\caption{Equipment Cost}
\begin{center}
\begin{tabular}{|c|c|c|c|c|}
\hline
\textbf{\textit{Name}} & \textbf{\textit{Unit}} & \textbf{\textit{Unit Price}} & \textbf{\textit{\makecell{Total \\ Price (BDT)}}} & \textbf{\textit{\makecell{Total \\ Price (USD)}}}\\
\hline
ESP32 & 3 & 450 & 1350 & 11.07 \\
\hline
Arduino Uno & 1 & 750 & 750 & 6.15 \\
\hline
L298N & 1 & 135 & 135 & 1.11 \\
\hline
1600mAH Battery & 1 & 1800 & 1800 & 14.75 \\
\hline
ECG Sensor & 1 & 750 & 750 & 6.15 \\
\hline
MAX30102 & 1 & 380 & 380 & 3.11 \\
\hline
MLX90614 & 1 & 995 & 995 & 8.16 \\
\hline
Sonar Sensor & 1 & 100 & 100 & 0.82 \\
\hline
Buck Converter & 1 & 75 & 75 & 0.61 \\
\hline
Gyroscope & 1 & 200 & 200 & 1.64 \\
\hline
Capacitor & 1 & 25 & 25 & 0.20 \\
\hline
Chassis Frame & 1 & 1200 & 1200 & 9.84 \\
\hline
Hand Gloves & 1 & 1200 & 1200 & 9.84 \\
\hline
Camera & 2 & 800 & 1600 & 13.11 \\
\hline
Raspberry Pi 5 & 1 & 14000 & 14000 & 114.75 \\
\hline
\textbf{Total} & & & \textbf{24610} & \textbf{201.72} \\
\hline
\end{tabular}
\label{tbl:cost}
\end{center}
\end{table}
\section{Result Analysis and Performance Measurement}
The smart wheelchair system was tested across modules—gesture navigation, ultrasonic and AI obstacle detection, physiological monitoring, emergency alerts, and web-based control—under real-world conditions to evaluate accuracy, responsiveness, and safety.
\subsection{Gesture-Based Wheelchair Control}
\begin{table}[h]
\scriptsize
\caption{Gesture Control Experiment Results}
\centering
\label{tab:gesture_results}
\begin{tabular}{|c|c|c|c|c|}
\hline
\textbf{\textit{Gesture}} & \textbf{\textit{Total Trials}} & \textbf{\textit{\makecell{Successful \\Movement}}} & \textbf{\textit{Failures}} & \textbf{\textit{\makecell{Success \\Rate (\%)}}} \\
\hline
Forward & 100 & 96 & 4 & 96\% \\
\hline
Backward & 100 & 98 & 2 & 98\% \\
\hline
Left & 100 & 93 & 7 & 93\% \\
\hline
Right & 100 & 95 & 5 & 95\% \\
\hline
\textbf{Total / Avg.} & \textbf{400} & \textbf{382} & \textbf{18} & \textbf{95.5\%} \\
\hline
\end{tabular}
\label{tab:gesture-results}
\end{table}
The wheelchair’s directional movement is controlled through hand gestures. A total of 400 trials were conducted across four gestures—forward, backward, left, and right. As shown in Table~\ref{tab:gesture_results}, the system achieved an overall success rate of 95.5\%, with highest accuracy for the backward gesture and slightly lower performance for the left turn due to a few misclassifications.
\subsection{Obstacle Detection Using Ultrasonic Sensor}
To ensure collision avoidance, the obstacle detection system was tested within a 20 cm range. The ultrasonic sensor achieved a 94\% success rate, with 47 successful detections out of 50 total trials and only 3 missed detections.
\subsection{YOLOv8-Based Object Detection Performance}
\begin{figure}[h]
\scriptsize
\centering
\begin{subfigure}[b]{0.49\linewidth}
    \centering
    \includegraphics[width=\linewidth]{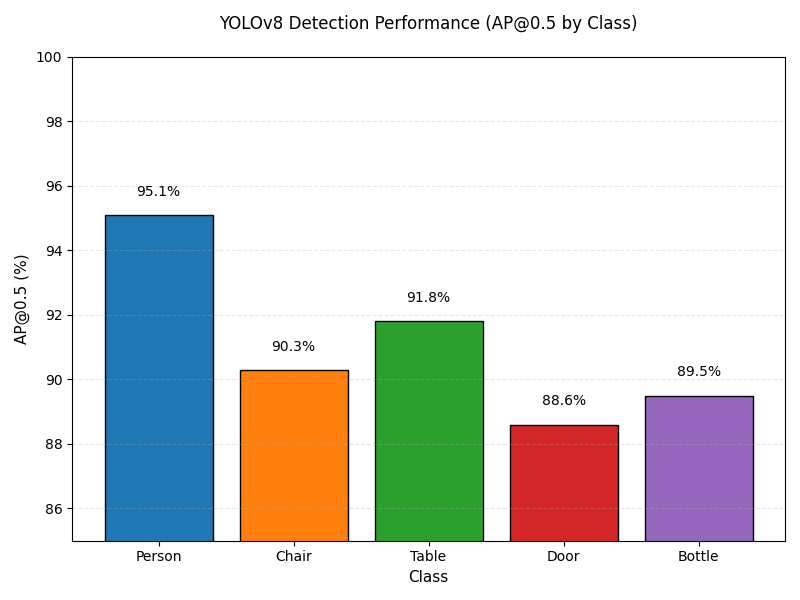}
    \caption{}
    \label{fig:detection_perf}
\end{subfigure}
\begin{subfigure}[b]{0.49\linewidth}
    \centering
    \includegraphics[width=\linewidth]{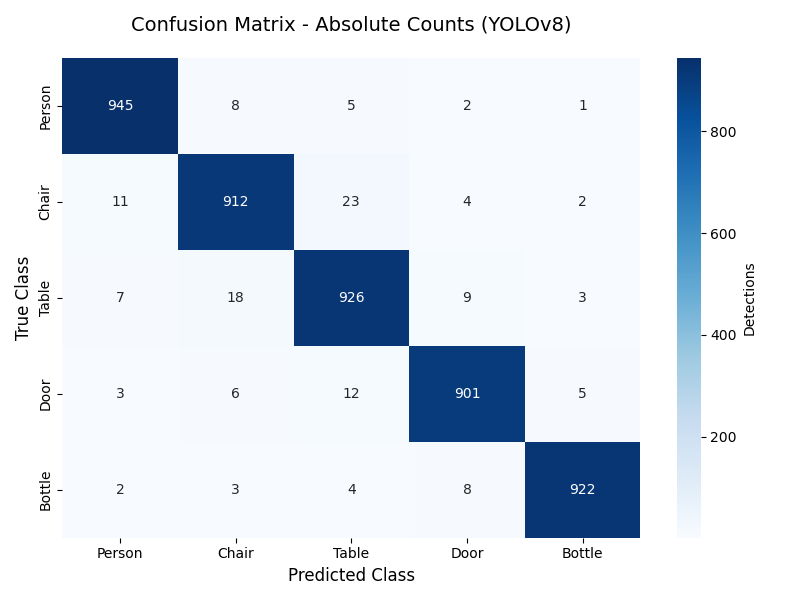}
    \caption{}
    \label{fig:confmatrix}
\end{subfigure}
\caption{YOLOv8 Evaluation Results: (a) Detection Performance of Test Classes (b) Confusion Matrix}
\label{fig:yolo_analysis}
\end{figure}

\begin{figure}[h]
\centering
\includegraphics[width=0.9\linewidth]{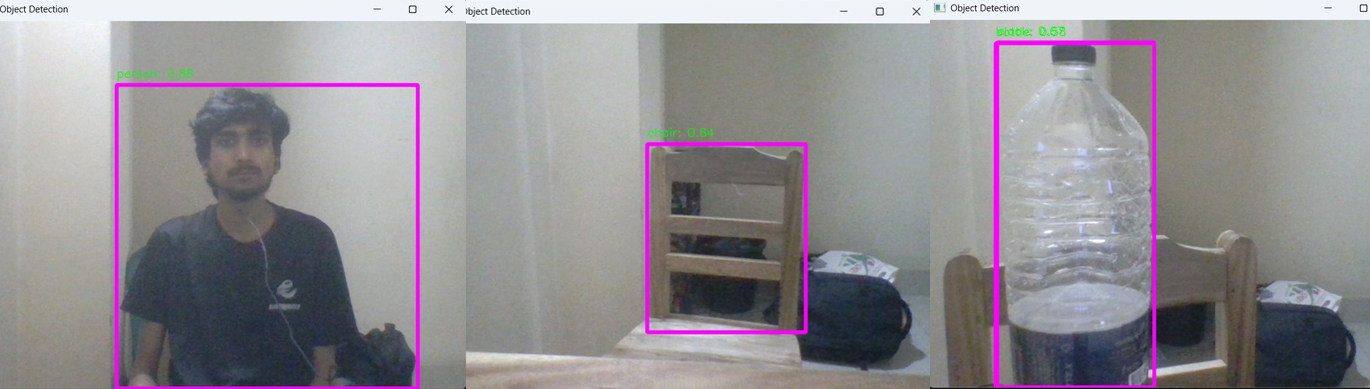}
\caption{Detection Output from YOLOv8 Inference}
\label{fig:yolo_output}
\end{figure}

For advanced visual navigation and obstacle identification, the YOLOv8 model was trained and tested on a subset of the COCO dataset \cite{2425}, focusing on five common indoor object classes: Person, Chair, Table, Door, and Bottle. The YOLOv8s model was utilized without architectural modifications to the backbone or detection head, prioritizing inference speed on the Raspberry Pi edge device. It was fine-tuned for 100 epochs on our custom dataset, which consisted of approximately 2,000 images (1,600 for training, 400 for validation) extracted and annotated from the COCO dataset. The training used a Stochastic Gradient Descent (SGD) optimizer with an initial learning rate of 0.01 and a batch size of 16.
The model demonstrated strong performance with Precision = 91.5\%, Recall = 90.2\%, and F1-score = 90.8\%. A detailed class-wise mAP@0.5 detection performance distribution illustrated in Figure \ref{fig:detection_perf} and confusion matrix is shown in Figure \ref{fig:confmatrix}. While only five classes were used in this evaluation, the model retains the capability to detect all standard COCO dataset objects in practical deployment scenarios. A sample output from the detection process is presented in Figure~\ref{fig:yolo_output} to visually demonstrate real-world model performance.

\subsection{Health Monitoring and Emergency Alert}
\begin{figure}[h]
\centering
\includegraphics[width=0.9\linewidth]{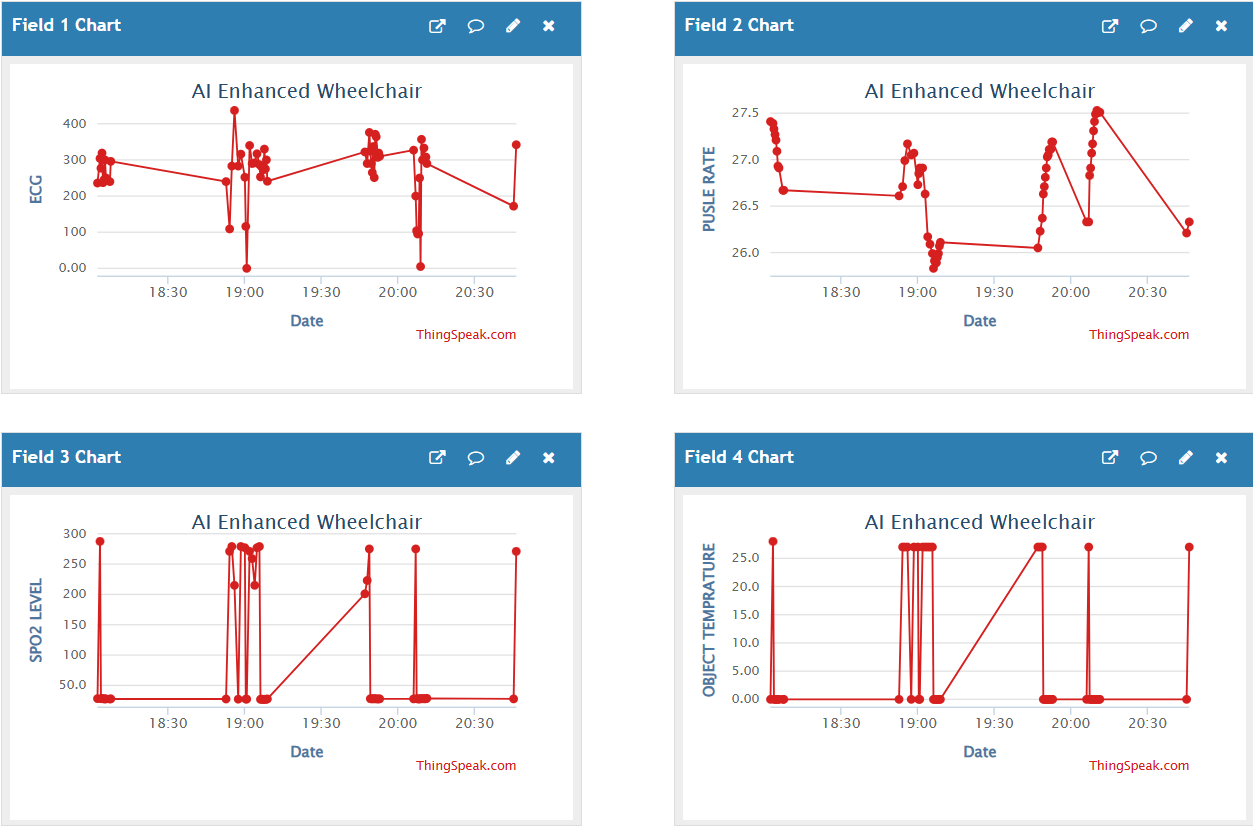}
\caption{ThingSpeak web dashboard showing live health monitoring.}
\label{fig:thingspeak_dashboard}
\end{figure}

To ensure user safety, the system includes onboard biomedical sensors for real-time monitoring of physiological parameters such as heart rate, body temperature, SpO$_2$, and ECG signals. These values are transmitted over Wi-Fi using the ESP32 microcontroller and visualized on the ThingSpeak IoT platform, shown in Figure~\ref{fig:thingspeak_dashboard}. Line charts for each parameter plot real-time data, allowing remote caregivers to monitor health continuously.
To prevent emergencies, the system checks for anomalies using pre-defined thresholds (e.g., temperature $> 100^\circ\mathrm{F}$ or SpO$_2 < 90\%$). Upon detecting abnormal readings, it automatically sends an emergency email alert, as shown in Figure~\ref{fig:email_alert}, including timestamp and sensor data.
\subsection{Remote Control via Web Dashboard}
\begin{figure}[h]
\centering
\begin{subfigure}{0.25\textwidth} 
    \includegraphics[width=\linewidth, height=3.5cm]{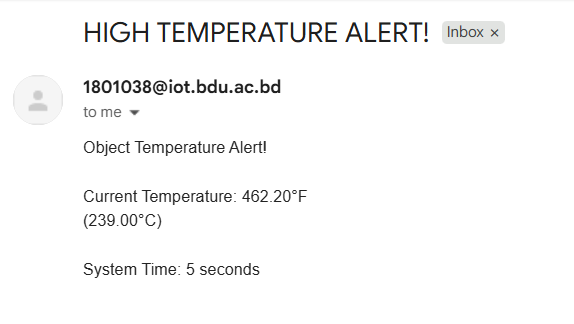}
    \caption{Emergency notification systems in Email}
    \label{fig:email_alert}
\end{subfigure}
\hfill 
\begin{subfigure}{0.23\textwidth} 
    \includegraphics[width=\linewidth, height=4cm]{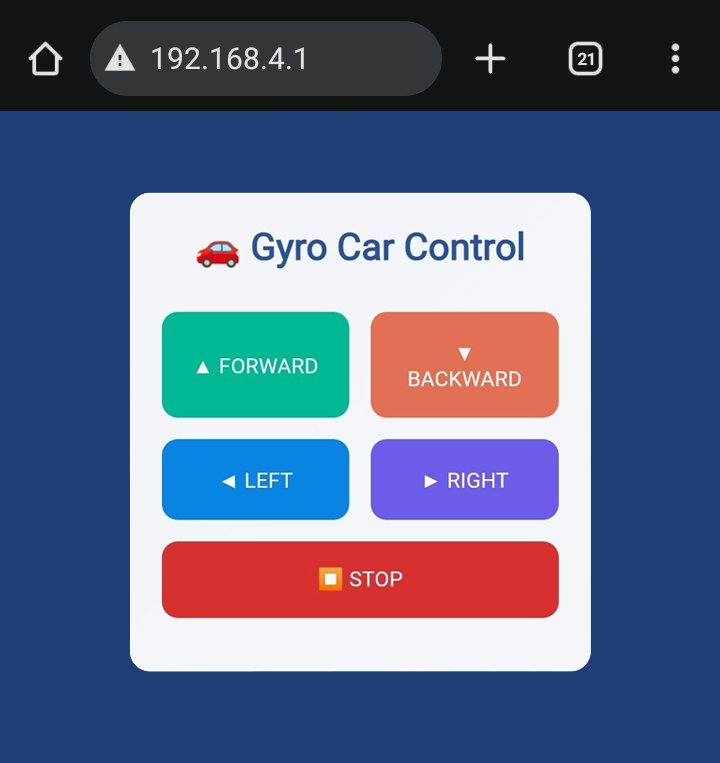}
    \caption{Remote Web Dashboard Control Interface}
    \label{fig:control}
\end{subfigure}
\caption{Combined figure showing notification and control interfaces}
\label{fig:combined}
\end{figure}
In addition to gesture-based operation, the wheelchair supports remote control via a web dashboard shown in Figure \ref{fig:control}. A secured web interface communicates with the onboard microcontroller over Wi-Fi, allowing a caregiver or remote user to control the wheelchair’s movement in real-time. Directional commands such as Forward, Backward, Left, and Right were tested via the dashboard, with low-latency execution observed in all trials.

\section{Discussion And Limitations}
The proposed smart wheelchair integrates gesture control, real-time object detection, and health monitoring at a low material cost of approximately \$201.72 (24,610 BDT). While the system performs reliably in controlled settings, several limitations warrant discussion for future improvement.
Gesture recognition accuracy can diminish with rapid or non-standard hand movements, as the current threshold-based algorithm lacks adaptive learning capabilities. The camera's field of view limits the effective range of object detection, and performance may degrade in low-light conditions or with small obstacles. The system's reliance on a stable Wi-Fi connection for module communication and cloud data upload poses a constraint for operation in areas with poor network coverage. Accurate health data acquisition is contingent on consistent and proper sensor placement, which may be challenging for some users. The current evaluation, while demonstrating technical feasibility, lacked extensive testing with differently-abled or elderly users in diverse, real-world environments to fully assess comfort and practicality. Furthermore, the prototype is built on a chassis frame; full integration into a commercial-grade wheelchair would require an additional investment of roughly \$90 (7,500 BDT) for structural components and sturdier parts (see Table \ref{tbl:cost}).
Despite these limitations, Table \ref{tab:comparison} highlights the unique combination of features in our system—hand tilt gestures, dual-layer obstacle detection (ultrasonic + YOLOv8), and multi-parameter health monitoring—at a significantly lower cost compared to related work, thereby enhancing its accessibility and potential functionality.
\begin{table}[h]
\scriptsize
\centering
\caption{Comparative Analysis of Smart Wheelchair Systems}
\begin{tabular}{|p{1.4cm}|p{1.2cm}|p{1.1cm}|p{1.2cm}|p{1.8cm}|}
\hline
\textbf{Feature} & \textbf{\cite{6,19}} & \textbf{\cite{7,9}} & \textbf{\cite{12,13,18}} & \textbf{Proposed System} \\
\hline
Control Modality & sEMG, MPU & Voice & -- & Hand Tilt (MPU6050) \\
\hline
Obstacle Detection & -- & Ultrasonic & Vision-based (YOLO variants) & Ultrasonic (HC-SR04) + YOLOv8 (Vision) \\
\hline
Health Monitoring (Parameters) & -- & SpO$_2$ & Heart Rate & ECG, SpO$_2$, Temperature \\
\hline
Key Differentiator & User-independent classifiers & Integrated voice control & Advanced vision navigation & Multi-modal integration, Low cost, IoT health alerts \\
\hline
Estimated Cost & High & Medium & High & Low (\$201.72) \\
\hline
\end{tabular}
\label{tab:comparison}
\end{table}
\section{Conclusion}
This paper presented an affordable AI-IoT smart wheelchair integrating gesture control, real-time navigation assistance, and continuous health monitoring. The system enables hands-free operation through natural movements, with safety features including ultrasonic obstacle detection, YOLOv8-based computer vision, and auditory feedback. Remote health monitoring via cloud platforms provides real-time vital sign tracking and automated alerts. Testing demonstrated reliable performance with 95.5\% gesture recognition accuracy, 94\% ultrasonic detection rate, and 90.8\% F1-score for object detection. The modular design using off-the-shelf components achieved a total cost of \$201.72, enhancing accessibility. Future work will focus on: (1) adaptive gesture recognition using machine learning, (2) expanded YOLOv8 training for diverse environments, (3) Bluetooth fallback mechanisms to reduce Wi-Fi dependence, (4) enhanced IoT security protocols, and (5) user studies with differently-abled individuals. The ultimate goal is transitioning this prototype to a full-scale, practical assistive mobility solution.


\begin{thebibliography}{99}

\bibitem{1}
World Health Organization, ``Disability and health,'' 2023. [Online]. Available: \url{https://www.who.int/news-room/fact-sheets/detail/disability-and-health}. Accessed: 2025-08-06.

\bibitem{2425}
R. Varghese and M. Sambath, ``Yolov8: A novel object detection algorithm with enhanced performance and robustness,'' in \textit{2024 International Conference on Advances in Data Engineering and Intelligent Computing Systems (ADICS)}, 2024, pp. 1–6.

\bibitem{2}
L. M. Suryavanshi, A. J. Chandraraj, K. Lochan, and P. Nag, ``Hand Gesture-Controlled Wheeled Mobile Robot for Prospective Application as Smart Wheelchairs,'' in \textit{Smart Sensors Measurement and Instrumentation}, Singapore: Springer, 2023, pp. 371–382.

\bibitem{akib1}
A. Z. Md Jalal Uddin, M. Rukaiya Begum, A. S. M. Ahsanul Sarkar Akib, K. Islam, A. Hasib, A. Giri, and A. Shahi, 
``LungNet: An Interpretable Machine Learning Framework for Early Lung Cancer Detection Using Structured Clinical Data,'' 
in \textit{2025 IEEE 13th Conference on Systems, Process \& Control (ICSPC)}, 
Melaka, Malaysia, 2025, pp. 181-186, doi: 10.1109/ICSPC68261.2025.11326300.

\bibitem{edubot}
A. Giri, A. S. M. A. S. Akib, A. Z. M. J. Uddin, M. S. Rahman, A. Hasib, M. Khadgi, M. F. Ferdous, M. Giri, and B. Shahi, ``EduBot: A Low-Cost Multilingual AI Educational Robot for Inclusive and Scalable Learning,'' in \textit{2025 3rd International Conference on Artificial Intelligence, Blockchain, and Internet of Things (AIBThings)}, 2025, pp. 1–6.

\bibitem{5}
I. Fadelli, ``Smart robotic wheelchair offers enhanced autonomy and control,'' TechXplore (Science X Network), Feb. 2025. [Online]. Available: \url{https://techxplore.com/news/2025-02-smart-robotic-wheelchair-autonomy.html}. Accessed: 6 August 2025.

\bibitem{6}
H. Iqbal, J. Zheng, R. Chai, Y. Zhang, and S. Su, ``Electric powered wheelchair control using user-independent classification methods based on surface electromyography signals,'' \textit{Medical \& Biological Engineering \& Computing}, vol. 62, pp. 167–182, 2024.

\bibitem{9}
D. Newaskar, J. Jangale, B. Bhoi, and S. Nalgirkar, ``Voice Controlled Wheelchair Along with Health Monitoring,'' \textit{Journal of Instrumentation Technology \& Innovations}, vol. 15, no. 01, pp. 1–6, 2025.

\bibitem{10}
M. S. Sadi, M. Alotaibi, M. R. Islam, M. S. Islam, T. Alhmiedat, and Z. Bassfar, ``Finger-Gesture Controlled Wheelchair with Enabling IoT,'' \textit{Sensors}, vol. 22, no. 22, p. 8716, 2022.

\bibitem{11}
T.-H. Nguyen, B.-V. Ngo, and T.-N. Nguyen, ``Vision-Based Hand Gesture Recognition Using a YOLOv8n Model for the Navigation of a Smart Wheelchair,'' \textit{Electronics}, vol. 14, no. 4, p. 734, 2025.

\bibitem{15}
M. R. Huda, M. L. Ali, and M. S. Sadi, ``Developing a real-time hand-gesture recognition technique for wheelchair control,'' \textit{PLOS ONE}, vol. 20, no. 4, p. e0319996, 2025.

\bibitem{19}
D. Shivamma, M. Manjula, S. B. Prasad, N. V. Konakalla, M. E. Reddy, and M. K. Ali, ``Smart Wheelchair Navigation Using Gesture-Based Control and IoT,'' in \textit{2025 International Conference on Intelligent and Innovative Technologies in Computing, Electrical and Electronics (IITCEE)}, 2025, pp. 1–4.

\bibitem{8}
J. Arshad, M. A. Ashraf, H. M. Asim, N. Rasool, M. H. Jaffery, and S. I. Bhatti, ``Multi-Mode Electric Wheelchair with Health Monitoring and Posture Detection Using Machine Learning Techniques,'' \textit{Electronics}, vol. 12, no. 5, p. 1132, 2023.

\bibitem{16}
A. Zahid, B. Poudel, D. Scott, J. Scott, S. Crouter, W. Li, and S. Swaminathan, ``PulseRide: A Robotic Wheelchair for Personalized Exertion Control with Human-in-the-Loop Reinforcement Learning,'' 2025. 

\bibitem{fall}
A. Giri, A. Hasib, M. Islam, M. F. Tazim, M. S. Rahman, M. Khadgi, and A. S. M. A. S. Akib, 
``Real-Time Human Fall Detection Using YOLOv5 on Raspberry Pi: An Edge AI Solution for Smart Healthcare and Safety Monitoring,'' 
in \textit{Proceedings of Data Analytics and Management}, 
Cham: Springer Nature Switzerland, 2026, pp. 493-507.

\bibitem{13}
B. Khalili and A. W. Smyth, ``SOD-YOLOv8 – Enhancing YOLOv8 for Small Object Detection in Traffic Scenes,'' 2024. 

\bibitem{17}
S. Chatzidimitriadis, S. M. Bafti, and K. Sirlantzis, ``Non-Intrusive Head Movement Control for Powered Wheelchairs: A Vision-Based Approach,'' \textit{IEEE Access}, vol. 11, pp. 65663–65674, 2023.

\bibitem{18}
J. Du, S. Zhao, C. Shang, and Y. Chen, ``Applying Image Analysis to Build a Lightweight System for Blind Obstacles Detecting of Intelligent Wheelchairs,'' \textit{Electronics}, vol. 12, no. 21, p. 4472, 2023.

\bibitem{12}
M. R. u. Rahman, P. Simonetto, A. Polato, F. Pasti, L. Tonin, and S. Vascon, ``OpenNav: Efficient Open Vocabulary 3D Object Detection for Smart Wheelchair Navigation,'' 2024. 

\bibitem{bionic}
M. W. Alim, A. Giri, A. S. M. A. S. Akib, N. Uddin, M. Islam, M. E. Arafat, and S. A. Tahmid, ``Affordable Bionic Hands With Intuitive Control Through Forearm Muscle Signals,'' in \textit{2025 IEEE 4th International Conference on Computing and Machine Intelligence (ICMI)}, 2025, pp. 1–6.

\bibitem{14}
Y. Xu, Q. Wang, J. Lillie, V. Kamat, and C. Menassa, ``CoNav Chair: Design of a ROS-based Smart Wheelchair for Shared Control Navigation in the Built Environment,'' 2025. 

\bibitem{7}
Y. Han, L. Zhou, W. Jiang, Y. Wang, S. Qiu, and D. Wu, ``Intelligent wheelchair human–robot interactive system based on human posture recognition,'' \textit{Journal of Mechanical Science and Technology}, vol. 38, pp. 4353–4363, 2024.

\bibitem{akib2}
A. Giri, A. S. M. A. S. Akib, A. Hasib, A. Acharya, M. A. Prithibi, An-Nafew, R. H. Rahman, M. R. Hossain, and H. I. Chowdhury Taha, ``Design and Development of a Cost Effective and Modular CNC Plotter for Educational and Prototyping Applications,'' in \textit{2025 IEEE 4th International Conference on Computing and Machine Intelligence (ICMI)}, 2025, pp. 1–6.

\bibitem{20}
MDN Web Docs, ``The WebSocket API (WebSockets),'' 2025. [Online]. Available: \url{https://developer.mozilla.org/en-US/docs/Web/API/WebSockets_API}. Accessed: 2025-08-06.

\bibitem{21}
Phipps Electronics, ``MPU6050 Datasheet,'' 2025. [Online]. Available: \url{https://www.phippselectronics.com/support/mpu6050-datasheet/}. Accessed: 2025-08-06.

\bibitem{22}
How2Electronics, ``ECG Monitoring with AD8232 ECG Sensor \& Arduino,'' 2023. [Online]. Available: \url{https://how2electronics.com/ecg-monitoring-with-ad8232-ecg-sensor-arduino/}. Accessed: 2025-08-06.

\bibitem{25}
Analog Devices / Maxim Integrated, ``MAX30102: High‑Sensitivity Pulse Oximeter and Heart‑Rate Sensor for Wearable Health,'' Data sheet (Rev. 1), 2018. [Online]. Available: \url{https://www.analog.com/media/en/technical-documentation/data-sheets/max30102.pdf}. Accessed: 6 August 2025.

\end{thebibliography}
\end{document}